\documentclass[runningheads]{llncs}

\usepackage{eccv}

\usepackage{eccvabbrv}

\usepackage{graphicx}
\usepackage{booktabs}
\usepackage{array}
\usepackage{multirow}
\usepackage{colortbl}
\usepackage{xcolor}
\usepackage{wrapfig}
\usepackage{caption}

\usepackage[accsupp]{axessibility}  

\usepackage{hyperref}

\usepackage{orcidlink}

\begin{document}

\title{Trajectory-Level Continuous Action Representation for Robotic Manipulation} 

\titlerunning{CAT}

\author{Tong Yang\inst{1,3}\textsuperscript{\dag} \and
Jingkai Jia\inst{2}\textsuperscript{\dag} \and
Yuecheng Xu\inst{2}\textsuperscript{\dag} \and
Xueyao Chen\inst{1} \and
Chi Zhang\inst{3} \and 
Wenqiang Zhang\inst{1,2}\textsuperscript{\ddag}}

\authorrunning{T.~Yang et al.}

\institute{Shanghai Key Lab of Intelligent Information Processing, College of Computer Science and Artificial Intelligence, Fudan University, Shanghai, China \and
College of Intelligent Robotics and Advanced Manufacturing, Fudan University, Shanghai, China \and
TeleAI, China Telecom, Shanghai, China \\
\email{tongyang23@m.fudan.edu.cn, wqzhang@fudan.edu.cn}}

\maketitle

\begingroup
\renewcommand{\thefootnote}{}
\footnotetext{
\textsuperscript{\dag} Equal contribution. \quad
\textsuperscript{\ddag} Corresponding author.
}
\endgroup

\begin{abstract}
We propose CAT, a trajectory-level continuous action representation framework for robotic manipulation. 
Existing visuomotor systems often entangle action representation with control frequency or rely on fixed temporal parameterizations. This leads to representational redundancy at high sampling rates and limits the modeling of critical motion. CAT instead encodes action trajectories within a fixed real-time interval into a set of continuous latent tokens. To ensure temporal consistency across varying control frequencies, we further incorporate a frequency-aware positional encoding that establishs a shared temporal coordinate system. Trajectory-level regularization further stabilizes the latent representation. This approach prevents representation growth with timestep density and avoids reliance on predefined temporal parameterizations. Extensive system-level evaluations on LIBERO, MimicGen, and real-world long-horizon manipulation tasks demonstrate that CAT-based policies consistently outperform both competitive VQ-based and continuous visuomotor baselines under matched training settings. Across various model backbones and control frequencies, CAT consistently improves success rates. These results highlight the advantages of trajectory-level continuous action modeling for scalable robotic manipulation across varying control rates.
\end{abstract}

\section{Introduction}
\label{sec:intro}

Learning visuomotor policies requires modeling continuous action trajectories whose influence on task outcomes is often unevenly distributed over time. In many manipulation tasks, only certain trajectory segments substantially affect task success. These include contact transitions, switches between free motion and interaction, and brief corrective adjustments. Other portions of the trajectory often consist of relatively smooth and repetitive motion. This temporal imbalance poses a challenge for policy learning. When modeling capacity is spread across the entire trajectory, temporally redundant regions can dominate the learning signal and dilute gradients associated with shorter but more consequential segments~\cite{pertsch2025fast}. As a result, action representations that do not account for this imbalance may struggle to preserve information carried by localized but important action variations.

Control frequency further amplifies this limitation in action representation. Under timestep-level discretization~\cite{lee2024behavior,wang2025vq,gong2025carp}, representational structure is directly tied to the temporal grid. As sampling frequency increases, trajectory length grows proportionally, expanding the timestep-level representation. However, task-relevant trajectory variation does not necessarily increase at the same rate, introducing temporal redundancy in the representation. To avoid this frequ-ency-driven expansion, some approaches adopt predefined temporal reparameterizations~\cite{pertsch2025fast,lyu2025omnisat,zhong2026freqpolicy}. 
These methods represent trajectories using a fixed set of parameters. While this prevents representation size from scaling with sampling frequency, the representational structure is still constrained by the chosen temporal basis. 

To address these limitations, we introduce CAT, a trajectory-level continuous action representation framework. 
CAT encodes action trajectories over a fixed real-time interval into a compact set of continuous latent tokens. 
The number of tokens remains constant and does not scale with the temporal grid used for execution. 
CAT employs a frequency-aware positional encoding to normalize timestep indices by control frequency.
This places trajectories sampled at different control rates in a shared temporal space.
Frequency-dependent variations remain observable in this space and are reflected in the latent representation.
This design allows CAT to learn trajectory-level latent representations without predefined temporal parameterizations. To ensure a stable latent structure,
the representation is optimized using reconstruction and regularization objectives.
These latent tokens serve as the action representation for policy learning and integrate seamlessly with a flow-matching policy learner.

We conduct extensive experiments in both simulated and real-world environments. On standard benchmarks including LIBERO~\cite{liu2023libero} and MimicGen~\cite{mandlekar2023mimicgen}, we compare CAT-based policies with competitive visuomotor systems. These include VQ-based VLA/LLM-style approaches (e.g., FAST, VQ-VLA, and CARP) as well as continuous-control baselines. CAT achieves higher average success rates across tasks, demonstrating strong system-level performance. To evaluate behavior under varying control frequencies, we further conduct experiments on RoboTwin 2.0~\cite{chen2025robotwin}, using it as a controlled multi-frequency evaluation platform. Across different sampling rates, CAT maintains consistent performance and generally outperforms baseline representations under matched training settings. We also validate CAT on real-world long-horizon manipulation tasks, where it achieves higher success rates than baseline systems, establishing a clear empirical advantage in extended-horizon manipulation.

In summary, our contributions are as follows:
\begin{itemize}
\item We introduce CAT, a trajectory-level continuous action representation framework. 
CAT encodes action trajectories using a fixed number of continuous latent tokens across control sampling rates. The representation is learned without relying on predefined temporal parameterizations.

\item We develop a frequency-conditioned trajectory modeling architecture that incorporates control frequency as an input signal. This architecture establishes shared temporal coordinates across control rates while preserving control-rate variations in the representation.

\item Through extensive system-level evaluations on both simulated and real-world robotic manipulation tasks, we show that CAT-based policies outperform competitive VQ-based and continuous-control baselines. 
The gains persist under varying control frequencies and in long-horizon real-robot settings.
\end{itemize}

\section{Related Work}
\label{sec:rel}

\textbf{Vision-Language-Action Models.} By leveraging the capabilities of pre-trained VLMs~\cite{liu2023visual,karamcheti2024prismatic,marafioti2025smolvlm}, Vision-Language-Action (VLA) models~\cite{brohan2022rt, brohan2023rt, kim2024openvla, black2024pi_0, intelligence2025pi_,team2024octo} have emerged as a powerful paradigm for robotic manipulation, enabling the mapping of multimodal inputs into corresponding action outputs. Current approaches generally fall into two main paradigms: autoregressive models and diffusion-based models. Autoregressive models~\cite{brohan2022rt, brohan2023rt, team2024octo,kim2024openvla} predict actions step by step, typically representing actions as discrete tokens to leverage large language model architectures for sequential reasoning and policy generation. In contrast, diffusion-based approaches~\cite{chi2025diffusion,black2024pi_0, intelligence2025pi_} directly model the denoising process in continuous control spaces, generating temporally coherent action trajectories. Despite their architectural differences, both paradigms model actions at the timestep level, with representational structures closely tied to temporal discretization.

\noindent \textbf{Action Representation Learning.} 
Prior VLAs commonly convert continuous action trajectories into structured sequence representations, which can broadly serve either discrete autoregressive prediction or continuous generative modeling. For discrete or autoregressive VLAs, action tokenizers provide a common way to convert continuous control signals into discrete symbolic sequences. Existing approaches either rely on predefined trajectory parameterizations, such as DCT coefficients~\cite{pertsch2025fast} and B-spline control points~\cite{lyu2025omnisat}, or learn vector-quantized codebooks over action sequences~\cite{lee2024behavior, mete2024quest,wang2025vq, gong2025carp,liu2025faster,dong2026actioncodec}. However, they bind representational structure to a specific temporal design—either via fixed basis functions, control-point allocations, or timestep-level discretization—causing the representation length to increase with finer sampling or to be constrained by the chosen parameterization.
For continuous generative policies, such as diffusion- or flow-matching-based policies, avoid symbolic action tokens but typically predict timestep-aligned action chunks, keeping the representation coupled to temporal resolution. FreqPolicy~\cite{zhong2026freqpolicy} introduces continuous spectral-domain tokens, yet still relies on a predefined spectral basis. In contrast, CAT learns trajectory-level continuous latent tokens over a fixed real-time interval, with token dimensionality remaining constant across sampling rates. This enables scalable action modeling under varying control frequencies without timestep-level discretization or predefined temporal/spectral bases.

\section{Method}
\label{sec:method}

\subsection{Overview} 
CAT is designed to address structural dependencies introduced by predefined temporal designs in existing action representations. Prior approaches either allocate representational units according to the timestep grid or compress trajectories using fixed temporal parameterizations. 

CAT removes this dependence through three key design principles. \textbf{Structural Decoupling} encodes trajectories over a fixed real-time interval into a fixed set of continuous latent tokens independent of timestep density. \textbf{Frequency Alignment} normalizes timesteps to align trajectories sampled at different control rates within a shared temporal space. \textbf{Latent Stability} combines reconstruction and contrastive regularization to maintain an expressive and discriminative latent space under a fixed token budget.

\begin{figure*}[t]
  \centering
  \includegraphics[width=0.9\linewidth]{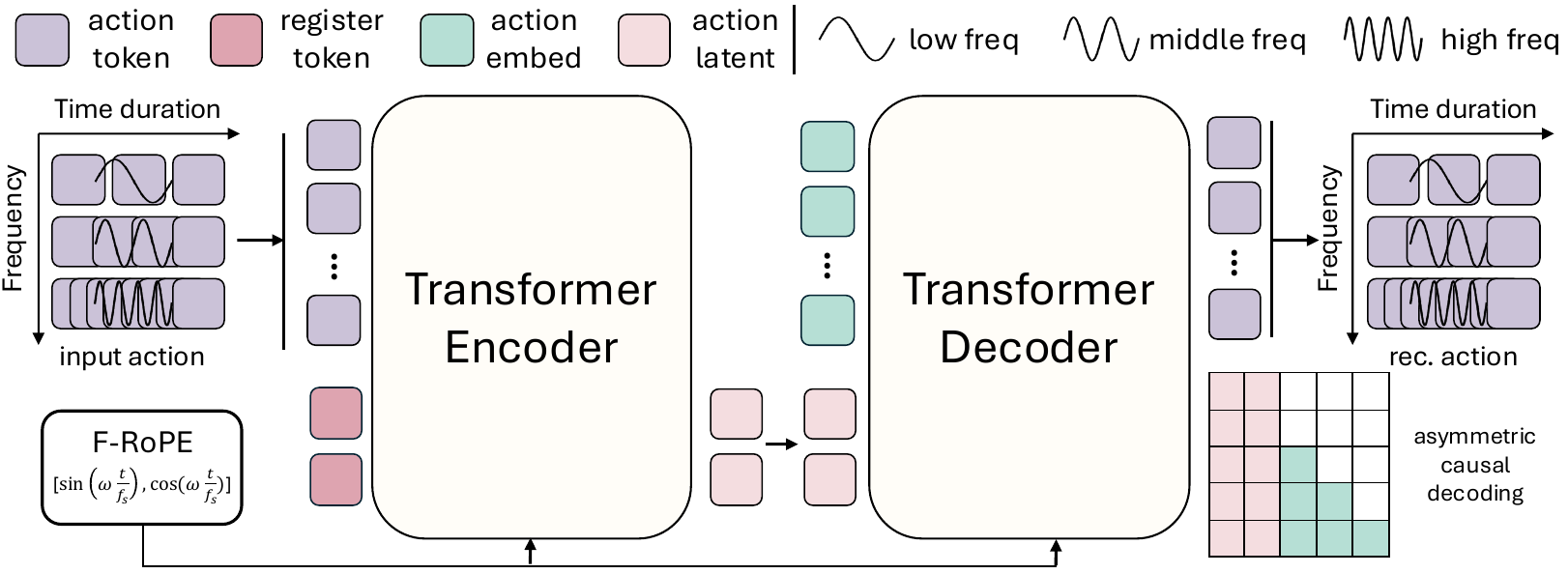}
  \caption{\textbf{Overview of CAT.}
    CAT performs trajectory-level continuous action modeling over action sequences defined within a fixed real-time window.
    A frequency-aware transformer encoder with F-RoPE injection maps each action sequence to a fixed set of continuous latent variables.
    An asymmetric causal decoder reconstructs the action sequence in a frequency-conditioned manner, enabling temporally consistent modeling across varying control rates.}
  \label{fig:overview}
\end{figure*}

As illustrated in Fig.~\ref{fig:overview}, CAT consists of a frequency-aware transformer encoder and decoder, both of which incorporate frequency-scaled timestep embeddings. The encoder compresses action trajectories into latent tokens, and the decoder reconstructs continuous trajectories. The learned latent tokens can be integrated into diffusion-based VLA frameworks as compact action representations for downstream generation.

\subsection{Trajectory-level Continuous Action Representation}

\textbf{Structural Decoupling:}
Existing action representations often derive their structure from a predefined temporal design. Under timestep-level discretization, representational units scale proportionally with the temporal grid. Alternatively, fixed temporal reparameterizations such as spectral bases or control-point schemes allocate parameters according to a predetermined basis. As a result, representational capacity becomes tied to the chosen temporal parameterization.

CAT removes this dependence by encoding trajectories within a fixed real-time window (T = 1 s) into a fixed number of continuous latent tokens.
Given an action sequence $\{a_t\}_{t=1}^{N}$ and their associated frequency-scaled timestep indices 
$\{t/f_s\}_{t=1}^{N}$, CAT learns the mapping:
\begin{equation}
\{z_i\}_{i=1}^K=f_{\theta}(\{a_t\}_{t=1}^{N}, \{t/f_s\}_{t=1}^{N}), \quad K \ll N .
\end{equation}
where $N = f_s T$ denotes the number of timesteps within the window under control frequency $f_s$,
and token count $K$ is fixed across sampling rates. The learnable register tokens 
$\{z_i\}_{i=1}^{K}$ attend to the full action sequence and aggregate 
trajectory-level information, ensuring that representational dimensionality does not scale with timestep density. The frequency-scaled timestep indices encode sampling-rate information and are used to construct frequency-aware positional embeddings for temporal alignment.

To reconstruct actions, CAT employs a transformer decoder that takes as input the register tokens concatenated with a set of learnable action embeddings $\{e_t\}_{t=1}^{N}$. Each action embedding $e_t$ corresponds to one timestep and is decoded into the reconstructed action $\hat a_t$.
The overall decoding process can be expressed in functional form as:
\begin{equation}
\{\hat a_t\}_{t=1}^{N}=g_{\phi}(\{z_i\}_{i=1}^K,\{e_t\}_{t=1}^{N}, \{t/f_s\}_{t=1}^{N})
\end{equation}
where $g_{\phi}$ is the transformer decoder parameterized by $\phi$. 
We design a decoder attention mask to regulate the information flow: all action embeddings can attend to the register tokens, while the action embeddings themselves follow causal attention along the temporal dimension. This asymmetric masking scheme enforces temporal causality during decoding, ensuring that action dependencies are modeled sequentially and improving the stability of generated trajectories.

However, structural decoupling alone does not resolve inconsistencies arising from different sampling densities. We therefore introduce an explicit frequency alignment mechanism.

\noindent \textbf{Frequency Alignment:} 
Under different control frequencies, the same timestep index corresponds to different physical times, making direct positional encoding inconsistent across control rates. 
To account for this discrepancy, CAT incorporates control frequency into positional encoding through a Frequency-aware Rotary Positional Embedding (F-RoPE). Each timestep index $t$ is scaled to $t/f_s$, producing time coordinates with spacing $1/f_s$ and making the positional encoding frequency-aware.
Formally, given a timestep $t$, its rotary embedding is defined in normalized time as:
\begin{equation}
\text{F-RoPE}\!\left(\frac{t}{f_s}\right)
= 
\big[
\sin(\omega_i \tfrac{t}{f_s}),
\;
\cos(\omega_i \tfrac{t}{f_s})
\big]_{i=1}^{d/2},
\end{equation}
where $\boldsymbol{\omega_i}$ denotes the $i$-th frequency coefficient, and $d$ is the embedding dimension.
These embeddings are then applied to the action query and action key vectors of the transformer layers in CAT. This embeds control-frequency information directly into the action representation, allowing trajectories sampled at different control rates to be distinguished.

\noindent \textbf{Latent Stability and Training Objective:} \label{sec:training} 
Encoding trajectories into a fixed number of latent tokens imposes a strong compression constraint: long action trajectories must be represented within a compact continuous latent space. Under this constraint, minimizing only reconstruction loss can cause distinct trajectories to occupy overlapping regions in latent space, reducing trajectory-level discriminability and weakening the stability of downstream generation.

To preserve discriminative structure within the fixed token budget, CAT combines reconstruction with trajectory-level contrastive regularization. Given an input action sequence $\{a_t\}_{t=1}^{N}$ and the decoder reconstruction $\{\hat{a}_t\}_{t=1}^{N}$, we use a reconstruction loss that penalizes discrepancies between the original and generated action trajectories:
\begin{equation}
\mathcal{L}_{\text{rec}}=\frac{1}{N} \sum_{t=1}^{N}||a_t-\hat{a}_{t}||_2^2.
\end{equation}

However, reconstruction alone does not explicitly regulate the global structure of the latent space. As a result, semantically distinct trajectories may become insufficiently separated in the latent representation. To encourage discriminative trajectory-level embeddings under a fixed representational budget, we introduce a contrastive regularization term. For each trajectory, we first obtain its embedding $\bar{z}$ by concatenating all register tokens,  $\bar{z}=[z_1||z_2||\cdots||z_K]$. 
The contrastive regularization is formulated as:
\begin{equation}
\mathcal{L}_{\text{reg}}
=
\frac{1}{B}
\sum_{i=1}^{B}
\log
\sum_{\substack{j=1 \\ j \neq i}}^{B}
\exp\!\left(
-\frac{\mathcal{D}(\bar{z}_i,\bar{z}_j)}{\eta}
\right)
\end{equation}
where $B$ denotes the mini-batch size, $i$ indexes the anchor trajectory embedding, and $j$ indexes the remaining samples within the same mini-batch. The function$\mathcal{D}(\cdot,\cdot)$ is a distance metric (e.g., cosine or Euclidean distance),
and $\eta$ is a temperature hyperparameter controlling the dispersion strength.
This objective penalizes small inter-trajectory distances, ensuring that trajectories with different motion semantics occupy distinct regions in the latent space while maintaining intra-trajectory cohesion enforced by $\mathcal{L}_{\text{rec}}$.

The overall training objective is formulated as:
\begin{equation}
\mathcal{L}_{\text{CAT}}= \mathcal{L}_{\text{rec}}+ \lambda_{\text{reg}}\mathcal{L}_{\text{reg}}
\end{equation}
where $\lambda_{\text{reg}}$ balances reconstruction accuracy and trajectory-level contrastive learning. Together, these two terms encourage CAT to learn a compact yet discriminative continuous action representation.

\begin{figure}[t]
\centering
\begin{minipage}{0.5\linewidth}
    \centering
    \includegraphics[width=\linewidth]{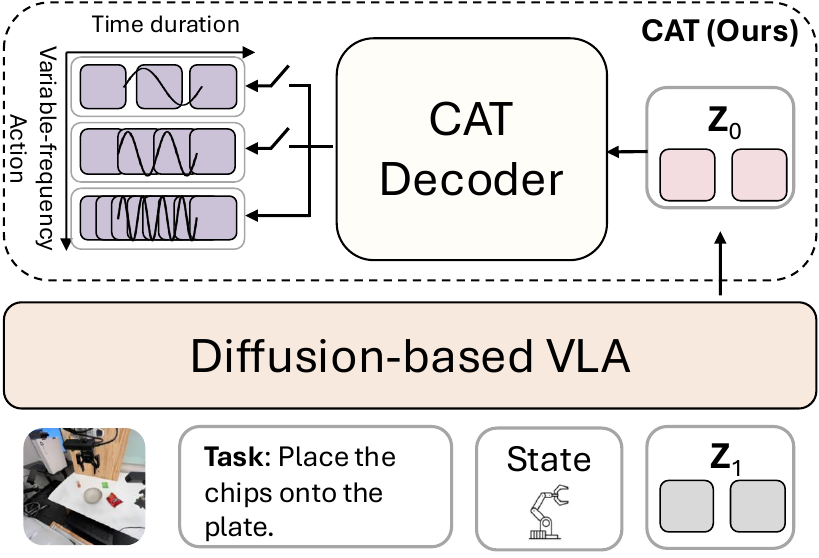}
\end{minipage}
\hfill
\begin{minipage}{0.45\linewidth}
\small
\textbf{Figure 2. Integration of CAT into a diffusion-based VLA framework.}
CAT replaces the raw action space with a compact continuous latent space,
decoupling representational complexity from control frequency.
Control frequency is incorporated as a conditioning signal,
allowing CAT-based policies to operate consistently under varying
execution rates within diffusion-based VLA systems.
\end{minipage}
\refstepcounter{figure}\label{fig:vla}
\end{figure}

\subsection{Integration into Diffusion-Based VLA} To demonstrate compatibility with continuous generative policies, we integrate CAT into a diffusion-based VLA framework. Diffusion-based VLAs generate temporally coherent action trajectories through iterative denoising, making them a natural setting to evaluate trajectory-level latent representations. CAT replaces the timestep-level action parameterization with a fixed-length continuous latent representation while preserving the original policy architecture.

The diffusion model, parameterized by $\psi$, operates on the latent trajectories $\mathbf{Z}=\{z_i\}_{i=1}^K$ and learns a continuous-time flow between noisy $\mathbf{Z}_1$ and clean latents $\mathbf{Z}_0$. At an intermediate time $\tau \in [0,1]$, the latent state $\mathbf{Z}_\tau$ is defined as a linear interpolation between the clean and noisy endpoints: $\mathbf{Z}_\tau=(1-\tau)\mathbf{Z}_0+\tau\mathbf{Z}_1 $. The model predicts a time-dependent velocity field $v_\psi(\mathbf{Z}_\tau, \tau, \textbf{c})$, where \textbf{c} represents task conditions such as language or scene context. Following the flow matching formulation, the training objective is defined as:
\begin{equation}
\mathcal{L}_{\text{FM}} =
\mathbb{E}_{\tau,\, \mathbf{Z}_0,\, \mathbf{Z}_1}
\Big[
\big\|
v_{\psi}(\mathbf{Z}_\tau, \tau, \textbf{c})
-
(\mathbf{Z}_1 - \mathbf{Z}_0)
\big\|_2^2
\Big]
\end{equation}

During inference, as shown in \cref{fig:vla}, the diffusion model refines noisy latent trajectories into clean representations, which are decoded by CAT into continuous actions.
The frequency-aware design of CAT enables consistent decoding under varying control rates while maintaining a fixed latent representation.

\section{Experiments}
We design our experiments to test structural hypotheses about trajectory-level action representation. Specifically, we evaluate whether modeling actions as trajectory-level continuous latent representations—decoupled from timestep discretization and fixed temporal parameterizations—leads to measurable improvements in policy learning, frequency robustness, and real-world deployment.

We formulate three hypotheses:
\begin{itemize}
\item \textbf{H1: Trajectory-Level Advantage.} 
Replacing timestep-level or discretization action representations with a trajectory-level continuous latent representation improves policy performance under matched training settings.

\item \textbf{H2: Frequency Robustness.} 
A trajectory-level representation with fixed latent budget remains stable under varying control frequencies, mitigating performance sensitivity to execution resolution.

\item \textbf{H3: Long-Horizon Stability.} 
Trajectory-level continuous modeling improves temporal coherence and stability in long-horizon real-world manipulation tasks under identical training budgets.
\end{itemize}
We evaluate these hypotheses through controlled experiments in simulation and real-world settings, including comparisons against timestep-level continuous policies, discretization-dependent action representations, and fixed-basis trajectory parameterizations under matched conditions.

\subsection{Simulation Experiments}

We evaluate CAT in simulation on two widely used robotic manipulation benchmarks: LIBERO~\cite{liu2023libero} and MimicGen~\cite{mandlekar2023mimicgen}. These benchmarks cover diverse manipulation settings, including spatial tasks, goal-conditioned control, long-horizon sequential manipulation, and multi-task learning.

\subsubsection{Evaluation on the LIBERO Benchmark}
\begin{table*}[t]
\centering
\caption{\textbf{Comparison on LIBERO.} Methods are grouped according to their action generation paradigm (autoregressive vs.\ diffusion/flow-based). Avg.\ denotes average success rate (\%).}
\begin{tabular}{lccccc}
\toprule
Method & Spatial & Object & Goal & Long & Avg. \\
\midrule
\rowcolor{gray!15}
\multicolumn{6}{l}{\hspace{-.5em}
\textit{Autoregressive-based VLA}}   \\
Octo~\cite{team2024octo} & 78.9 & 85.7 & 84.6 & 51.1 & 75.1 \\
OpenVLA~\cite{kim2024openvla} & 84.7 & 88.4 & 79.2 & 53.7 & 75.9 \\
SmolVLA-FAST & 87.0 & 93.0 & 90.0 & 68.0 & 84.5 \\
$\pi_0$-FAST~\cite{pertsch2025fast} & 96.4 & 96.8 & 88.6 & 60.2 & 85.0 \\
SmolVLA-VQ-VLA & 90.0 & 94.0 & 86.0 & 71.0 & 85.3 \\
OmniSAT~\cite{lyu2025omnisat} & 94.1 & 98.7 & 94.6 & 86.0 & 93.4 \\
\midrule
\rowcolor{gray!15}
\multicolumn{6}{l}{\hspace{-.5em}
\textit{Diffusion/Flow-based VLA}}   \\
Diffusion Policy~\cite{chi2025diffusion} & 78.3 & 92.5 & 68.3 & 50.5 & 72.4 \\
4D-VLA~\cite{zhang20264d} & 93.8 & 92.8 & 95.6 & 86.5 & 92.2 \\
CogACT~\cite{li2024cogact} & 97.2 & 98.0 & 90.2 & 88.8 & 93.2 \\
\cmidrule(lr){1-6}
SmolVLA~\cite{shukor2025smolvla} & 90.0 & 96.0 & 92.0 & 71.0 & 87.3 \\
\textbf{SmolVLA-CAT (Ours)} & 95.0 & 97.0 & 94.0 & 77.0 & 90.8 \\

\cmidrule(lr){1-6}
$\pi_0$~\cite{black2024pi_0} & 96.8 & 98.8 & \textbf{95.8} & 85.2 & 94.2 \\
\textbf{$\pi_0$-CAT (Ours)} & \textbf{98.2} & \textbf{99.8} & 95.6 & \textbf{92.4} & \textbf{96.5} \\
\bottomrule
\end{tabular}
\label{tab:libero-comparison}
\end{table*}
\paragraph{Setup.}
We evaluate CAT on the LIBERO benchmark~\cite{liu2023libero}, which contains four task suites: Spatial, Object, Goal, and Long.

For controlled comparisons, we construct four SmolVLA-based variants that share the same backbone architecture, policy network structure, dataset, preprocessing, and training schedule: 
(i) the original SmolVLA baseline with timestep-level continuous action prediction; 
(ii) SmolVLA-FAST and SmolVLA-VQ-VLA, which adopt discretization-dependent trajectory representations with autoregressive token prediction; and 
(iii) SmolVLA-CAT, which models each action chunk using continuous trajectory-level latent tokens. 
Across these variants, the network architecture remains identical; the only differences lie in the action representation and the corresponding prediction objective.

To evaluate cross-model generality, we further replace the action representation of the larger diffusion-based model $\pi_0$~\cite{black2024pi_0} with CAT while keeping its backbone architecture, optimizer, and training protocol unchanged.

We also report representative autoregressive and diffusion-based VLA systems for context. These methods may differ in backbone scale and training data and are not intended as controlled comparisons.

\paragraph{Results.}
Table~\ref{tab:libero-comparison} reports the LIBERO results.
Under the SmolVLA backbone, replacing timestep-level continuous action prediction with CAT improves the average success rate from 87.3\% to 90.8\% (+3.5). This controlled comparison highlights the benefit of shifting from timestep-level modeling to trajectory-level continuous latent representation, suggesting that allocating modeling capacity at the trajectory level leads to more effective policy learning.

Under the same backbone and identical policy architecture, SmolVLA-CAT consistently outperforms SmolVLA-FAST and SmolVLA-VQ-VLA. 
The consistent gains suggest that modeling actions as continuous trajectory-level latents provides a more effective inductive bias than discretization-dependent trajectory summaries—such as predefined-basis or timestep-level representations—within the same architecture.

Moreover, integrating CAT into the larger diffusion-based model $\pi_0$ further improves performance from 94.2\% to 96.5\%, indicating that the representational gain generalizes across model scales.

Overall, these results show that modeling actions as continuous trajectory-level latents provides a more effective inductive bias for visuomotor policy learning. This leads to consistent improvements across different architectures and model scales.

\begin{table*}[t]
    \caption{\textbf{Comparison of success rates (\%) on 8 manipulation tasks in MimicGen~\cite{mandlekar2023mimicgen}.} Bold numbers indicate the best performance.}
    \centering
    \resizebox{0.9\textwidth}{!}{%
    \begin{tabular}{lccccccccc}
    \toprule
    Policy
    & Coffee & Hammer & Mug & Nut 
    & Square & Stack & Stack 3 
    & Threading & Avg.   \\
    \midrule
    TCD~\cite{liang2024skilldiffuser}    & 77 & 92 & 53 & 44 & 63 & 95 & 62 & 56 & 67.3 \\
    SDP~\cite{wang2024sparse}    & 82 & \textbf{100} & 62 & 54 & 82 & 96 & 80 & 70 & 78.3 \\
    CARP~\cite{gong2025carp}   & 86 & 98 & \textbf{74} & 78 & 90 & \textbf{100} & 82 & 70 & 84.8  \\
    \midrule
    DP ~\cite{chi2025diffusion}   & 89 & 99 & 67 & \textbf{84} & 90 & 99 & 76 & \textbf{82} & 85.8  \\
    \textbf{DP-CAT (Ours)}   & \textbf{91} & 99 & 68 & 81 & \textbf{91} & 99 & \textbf{85} & 81 & \textbf{86.9}  \\
    \bottomrule
    \end{tabular}
    
    }
     \label{tab:multi-task-eval}
\end{table*}

\subsubsection{Evaluation on the MimicGen Benchmark}

\paragraph{Setup.}
We evaluate CAT on MimicGen~\cite{mandlekar2023mimicgen}, a large-scale imitation learning benchmark built on Robomimic with diverse manipulation tasks and broad initial-state distributions. Following prior multi-task settings~\cite{liang2024skilldiffuser, gong2025carp, wang2024sparse}, we train a single policy jointly on 8 robosuite tasks.

We compare five representative methods: 
task conditioned diffusion (TCD)~\cite{liang2024skilldiffuser} and sparse diffusion policy (SDP)~\cite{wang2024sparse} as prior baselines, 
CARP~\cite{gong2025carp} as an autoregressive multi-task policy with VQ-based action discretization, 
a flow-matching Diffusion Policy (DP) that replaces the original diffusion objective with flow matching~\cite{lipman2022flow}, 
and DP-CAT, which substitutes timestep-level action prediction with trajectory-level continuous latent modeling.

For fair comparison, DP matches the parameter count of CARP. All models are trained under identical data splits and evaluation protocols. We report average success rates over 50 rollouts per task.

\paragraph{Results.}

Table~\ref{tab:multi-task-eval} summarizes performance across 8 MimicGen tasks. 
The flow-matching Diffusion Policy (DP) achieves an average success rate of 85.8\%, while integrating CAT improves performance to 86.9\% under matched parameter counts (+1.1). This indicates that trajectory-level continuous latent modeling provides gains over timestep-level prediction in a shared multi-task architecture.

The largest improvement appears on the Stack 3 task, where DP-CAT achieves 85\% compared to 76\% for DP (+9). Stack 3 involves sequential multi-stage stacking and longer effective control horizons. The observed gain suggests that trajectory-level representation remains effective in extended sequential settings.

\subsubsection{Control-Frequency Evaluation on RoboTwin~2.0}

\paragraph{Setup.}
To evaluate CAT under varying control frequencies, we repurpose RoboTwin 2.0~\cite{chen2025robotwin} as a controlled testbed for frequency analysis. Although RoboTwin is not originally designed as a control-frequency benchmark, its simulation framework enables systematic variation of control rates while keeping task configurations fixed. Experiments are conducted under the Easy split, as our focus is representation–frequency decoupling rather than scene generalization.

We collect manipulation trajectories at four control frequencies: 10Hz, 16.7Hz, 25Hz, and 50Hz. 
All trajectories are represented in the delta joint format to maintain a unified action space across frequencies.
Using the aggregated multi-frequency dataset, we first train CAT as a shared action representation model. 
Based on the learned representation, we then train a separate SmolVLA-CAT policy for each control frequency, following the same backbone architecture, optimizer, and training protocol as the corresponding baseline SmolVLA model. 
Evaluation is conducted on a subset of 8 manipulation tasks from RoboTwin~2.0, with 100 rollouts per task. We report average success rates across tasks.

\begin{table*}[t]
\centering
\caption{\textbf{Success rates (\%) on 8 RoboTwin~2.0 manipulation tasks evaluated under different control frequencies.} Results are averaged over 100 rollouts per task. Higher values between SmolVLA (Baseline) and SmolVLA-CAT (Ours) are highlighted in bold.}
\resizebox{0.98\linewidth}{!}{
\begin{tabular}{lcccccccccc}
\toprule
\multirow{2}{*}{Task} & 
\multicolumn{2}{c}{10Hz} & 
\multicolumn{2}{c}{16.7Hz} & 
\multicolumn{2}{c}{25Hz} & 
\multicolumn{2}{c}{50Hz} &
\multicolumn{2}{c}{Avg.(Task)} \\
\cmidrule(r){2-3} \cmidrule(r){4-5} \cmidrule(r){6-7} \cmidrule(r){8-9} \cmidrule(r){10-11}
 & Baseline & Ours & Baseline & Ours & Baseline & Ours & Baseline & Ours & Baseline & Ours \\
\midrule
Adjust Bottle         & \textbf{76} & 68 & 77 & \textbf{78} & 69 & \textbf{72} & 78 & \textbf{86} & 75.0 & \textbf{76.0} \\
Click Alarmclock      & \textbf{54} & 47 & 52 & \textbf{54} & 51 & \textbf{58} & 43 & \textbf{50} & 50.0 & \textbf{52.3} \\
Grab Roller           & 17 & \textbf{47} & 46 & \textbf{66} & 49 & \textbf{58} & 56 & \textbf{71} & 42.0 & \textbf{60.5} \\
Lift Pot              & 13 & \textbf{25} & 26 & \textbf{28} & 28 & \textbf{29} & 20 & \textbf{27} & 21.8 & \textbf{27.3} \\
Place Burger Fries    & \textbf{38} & 37 & 24 & \textbf{26} & \textbf{29} & \textbf{29} & 26 & \textbf{31} & 29.3 & \textbf{30.8} \\
Place Container Plate & 43 & \textbf{71} & \textbf{64} & 57 & 50 & \textbf{76} & \textbf{46} & 40 & 50.8 & \textbf{61.0} \\
Press Stapler         & 33 & \textbf{52} & 36 & \textbf{52} & 19 & \textbf{50} & \textbf{50} & 48 & 34.5 & \textbf{50.5} \\
Rotate QRcode         & 8  & \textbf{18} & \textbf{12} & 8 & 8 & \textbf{12} & 1 & \textbf{9} & 7.3 & \textbf{11.8} \\
\midrule
Avg.(Frequency)                  & 35.3 & \textbf{45.6} & 42.1 & \textbf{46.1} & 37.9 & \textbf{48.1} & 40.0 & \textbf{45.3} & - & - \\
\bottomrule
\end{tabular}
}
\label{tab:freq_comparison}
\end{table*}

\paragraph{Results.}
Table~\ref{tab:freq_comparison} compares SmolVLA and SmolVLA-CAT under four control frequencies on RoboTwin~2.0. SmolVLA-CAT achieves higher average success rates than the baseline at all evaluated frequencies. Specifically, SmolVLA-CAT attains average success rates of 45.6\%, 46.1\%, 48.1\%, and 45.3\% at 10Hz, 16.7Hz, 25Hz, and 50Hz, respectively, compared to 35.3\%, 42.1\%, 37.9\%, and 40.0\% for SmolVLA. This corresponds to gains of +10.3, +4.0, +10.2, and +5.3 points. 

Beyond the average improvement, CAT outperforms the baseline in 24 out of 32 task–frequency pairs, indicating that the gain is broadly consistent rather than driven by a small number of outlier tasks. Averaged across frequencies, the largest task-level improvements appear on \textit{Grab Roller} (+18.5), \textit{Press Stapler} (+16.0), \textit{Place Container Plate} (+10.3), which involve short but critical interaction transitions. In contrast, improvements are smaller on smoother tasks such as \textit{Adjust Bottle} (+1.0) and \textit{Place Burger Fries} (+1.5). This pattern supports the motivation of CAT: trajectory-level representations are most beneficial when success depends on localized but important action variations.

Interestingly, the optimal control frequency varies across tasks. Some tasks, such as \textit{Grab Roller} and \textit{Adjust Bottle}, benefit from higher control rates, while others such as \textit{Place Container Plate} achieve their best performance at lower frequencies. This observation suggests that the temporal resolution required for effective control depends on task dynamics and interaction complexity.

\subsection{Real-World Long-Horizon Manipulation}

\begin{figure}[t]
  \centering
  \includegraphics[width=\linewidth]{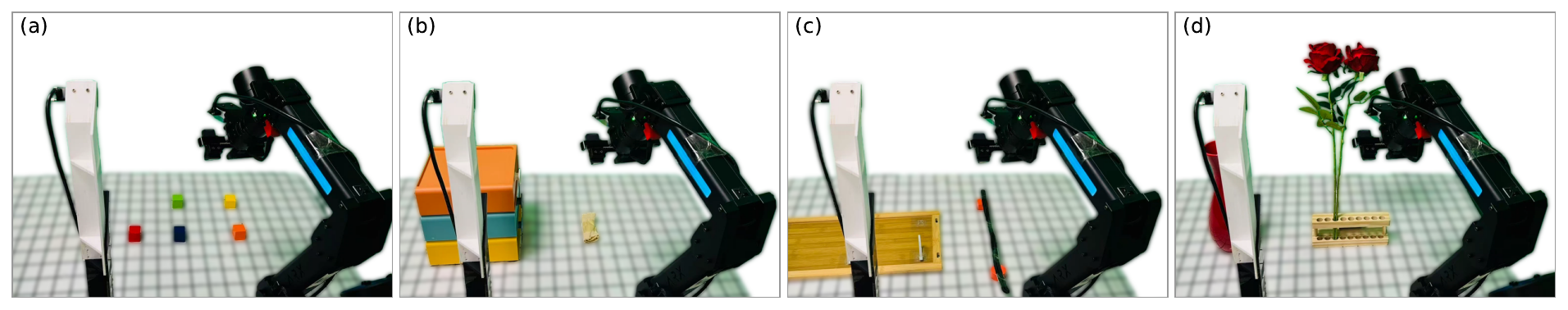}
  \caption{\textbf{Examples of initial states for 4 real-world tasks.} Each panel shows the third-person camera, the robot arm with the wrist camera mounted, and the manipulated objects. Tasks: (a) Stack-5, (b) Drawer, (c) Flower, (d) Rope.}
  \label{fig:real-world}
\end{figure}

\begin{table}[t]
    \caption{\textbf{Normalized task score (\%) on 4 real-world long-horizon tasks.}
    Baseline denotes SmolVLA, and Ours denotes SmolVLA integrated with CAT (SmolVLA-CAT). Bold numbers indicate the best performance.}
    \centering
    \begin{tabular}{lccccc}
        \toprule
        Models & Stack-5 & Drawer & Rope & Flower & Avg. \\
        \midrule
        Baseline     & 29 & 49 & 62 & 40 & 45.0 \\
        Ours & \textbf{43} & \textbf{61} & \textbf{70} & \textbf{68} & \textbf{60.5} \\
        \bottomrule
    \end{tabular}
    \label{tab:real-world-result}
    \vspace{-5mm}
\end{table}

\begin{figure*}[t]
\centering

\begin{subfigure}{\linewidth}
    \centering
    \includegraphics[width=\linewidth]{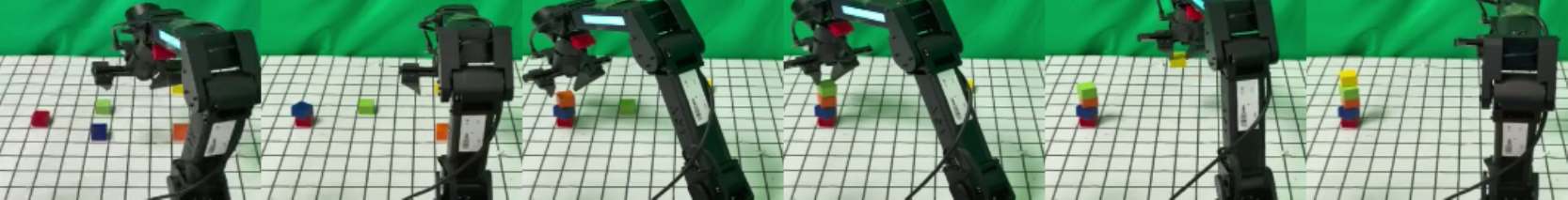}
    \caption{Stack-5: stack blocks in yellow-green-orange-blue-red order}
\end{subfigure}

\begin{subfigure}{\linewidth}
    \centering
    \includegraphics[width=\linewidth]{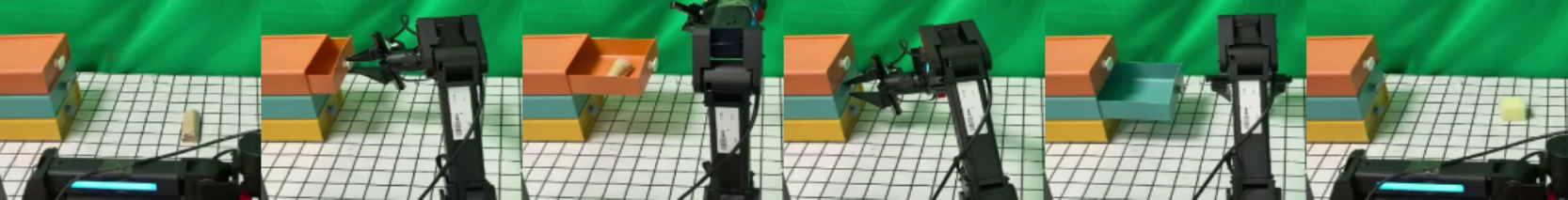}
    \caption{Drawer: open the top drawer, put the tissue into the top drawer and close drawer; open the middle drawer, take the sponge out and close the drawer.}
\end{subfigure}

\begin{subfigure}{\linewidth}
    \centering
    \includegraphics[width=\linewidth]{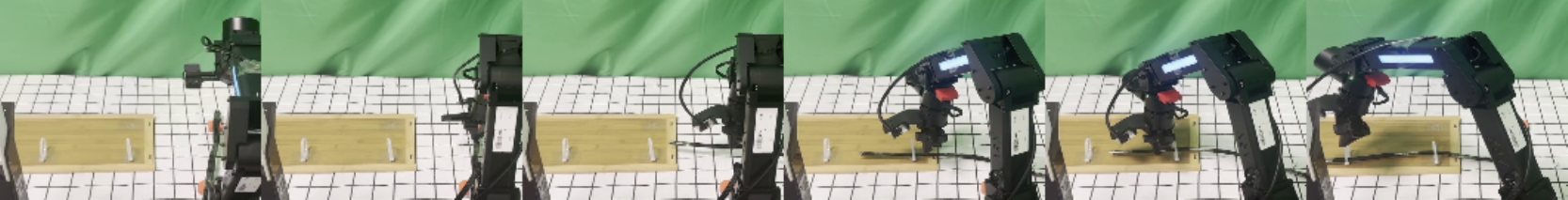}
    \caption{Rope: pick the rope, thread it into the first ring, drop it, pick it from the other side of the ring and thread it into the second ring.}
\end{subfigure}

\begin{subfigure}{\linewidth}
    \centering
    \includegraphics[width=\linewidth]{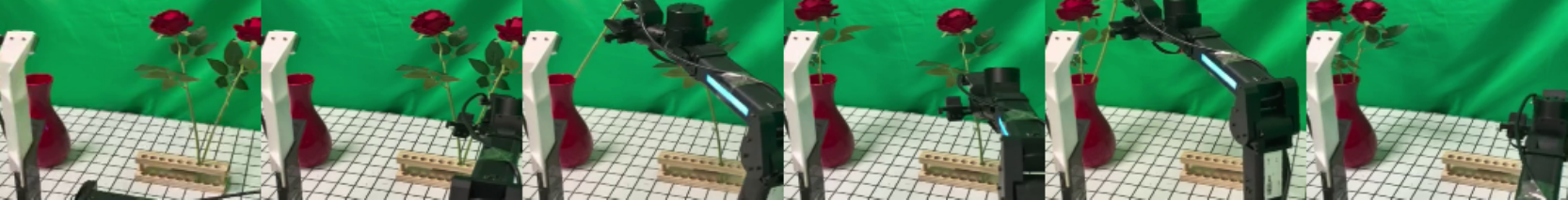}
    \caption{Flower: pick one flower and insert it into the vase; pick the flower left and insert it into the vase.}
\end{subfigure}

\caption{
\textbf{Qualitative results on real-world tasks.} Representative examples include tasks from Stack-5, Drawer, Rope, and Flower.
}
\label{fig:real-world-tasks}
\vspace{-5mm}
\end{figure*}

\paragraph{Setup.}
We evaluate CAT on four long-horizon real-world manipulation tasks (see Fig.~\ref{fig:real-world}): 
\textbf{Stack-5} (stack four cubes sequentially on a red base cube to form a five-cube tower), 
\textbf{Drawer} (open the top drawer, place a tissue inside, and close it; then open the middle drawer and remove a sponge), 
\textbf{Rope} (thread a rope through two rings sequentially), and 
\textbf{Flower} (pick and place two roses into a vase, one at a time). 
These tasks require sequential completion of multiple manipulation stages and precise control throughout execution. 
The experiments are conducted with a single ARX~R5 arm equipped with one third-person camera and one wrist camera.

For each task, we collect 100 teleoperated demonstrations and train a task-specific policy. 
As the baseline, SmolVLA~\cite{shukor2025smolvla} is trained for 180k steps on each task. 
For our method, CAT is first pretrained on the Open X-Embodiment dataset~\cite{collaboration2023open} following the VQ-VLA~\cite{wang2025vq} setup, then fine-tuned for 1k steps on the target task data, after which SmolVLA is trained using the CAT-based trajectory-level action representation for the same number of task-specific policy training steps as the baseline. 

For evaluation, each task is decomposed into 4--8 sub-steps according to its execution horizon. 
Each completed sub-step contributes one point, and the total is normalized to a 0--100 scale, where 100 indicates that all sub-steps are completed. 
We report the average normalized task score over 10 evaluation episodes for each task.

\paragraph{Results.}
Table~\ref{tab:real-world-result} reports the normalized task scores on the four real-world tasks. CAT improves the average score from 45.0 to 60.5, yielding a gain of 15.5 points over the baseline. 
It outperforms the baseline on all four tasks, with gains of +14 on Stack-5, +12 on Drawer, +8 on Rope, and +28 on Flower. These results show that trajectory-level continuous latent modeling remains effective in real-world manipulation.
All four tasks are long-horizon and require multiple sub-steps to be completed in sequence. The consistent improvements across stacking, articulated-object manipulation, and deformable-object manipulation suggest that CAT is particularly beneficial in real-world sequential manipulation scenarios where intermediate progress directly affects the final outcome.

Representative qualitative examples of successful real-world executions are shown in Fig~\ref{fig:real-world-tasks} for reference.

\subsection{Ablation studies}
To investigate the specific contributions of the key components in our proposed CAT, we conduct a series of ablation studies focusing on four major design factors: the positional encoding scheme, the number of registers $K$, the weight of the regularization loss $\lambda_{\text{reg}}$, and the decoder attention type. Unless otherwise specified, all ablations are performed on LIBERO under the same training and evaluation protocol as in the main experiments.

\paragraph{Positional Encoding.}
\begin{wrapfigure}{r}{0.44\linewidth}
  \vspace{-32pt}
  \centering
  \includegraphics[width=\linewidth]{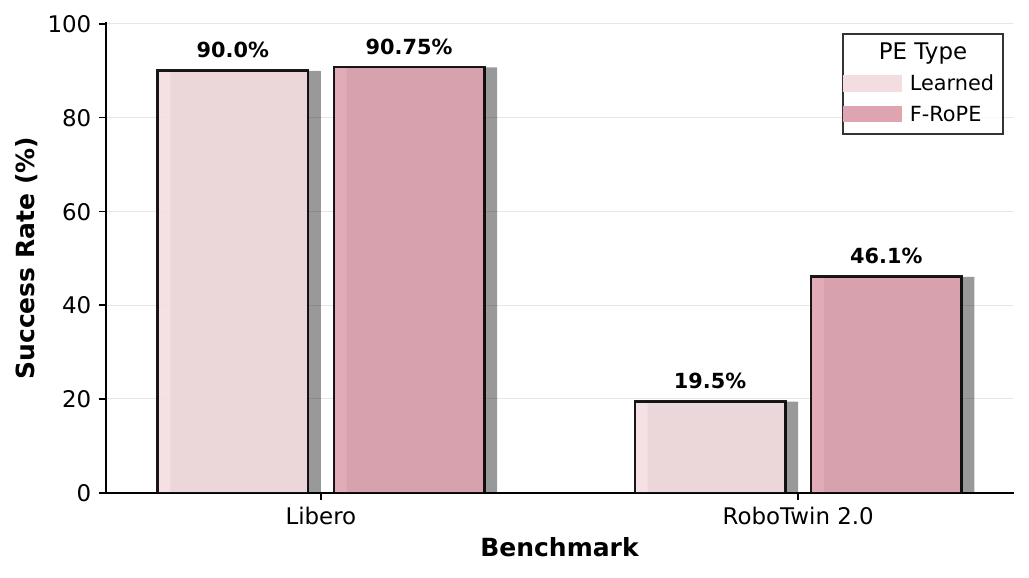}
  \vspace{-9mm}
  \caption{\textbf{PE comparison.} F-RoPE vs learned PE.}
  \label{fig:pe_comparison}
  \vspace{-30pt}
\end{wrapfigure}
We compare frequency-aware rotary positional embeddings (F-RoPE) with learned positional embeddings on both LIBERO and RoboTwin 2.0. As shown in \cref{fig:pe_comparison}, F-RoPE gives slightly better performance on LIBERO and larger gains on RoboTwin~2.0, where evaluation spans multiple control frequencies. These results suggest that F-RoPE is a more effective positional encoding choice, particularly when the data and evaluation involve varying sampling rates.

\begin{table}[t]
\centering
\caption{\textbf{Ablation studies on LIBERO.}
S: Spatial, O: Object, G: Goal, L: Long-horizon.
Performance of CAT under different design choices.}
\label{tab:ablation_all}

\resizebox{0.94\columnwidth}{!}{%
\begin{minipage}{\columnwidth}
\centering
\setlength{\tabcolsep}{3pt}
\renewcommand{\arraystretch}{1.05}
\small

\begin{subtable}[b]{0.32\linewidth}
\centering
\begin{tabular}{lccccc}
\toprule
$K$ & S & O & G & L & Avg \\
\midrule
2 & 91 & \textbf{98} & 91 & 72 & 88.0 \\
4 & \textbf{95} & 97 & \textbf{94} & \textbf{77} & \textbf{90.8} \\
6 & 92 & \textbf{98} & 93 & 74 & 89.3 \\
8 & 90 & 96 & 92 & 71 & 87.3 \\
\bottomrule
\end{tabular}
\caption{Number of Registers}
\label{tab:ablation-register}
\end{subtable}
\hfill
\begin{subtable}[b]{0.32\linewidth}
\centering
\begin{tabular}{lccccc}
\toprule
$\lambda$ & S & O & G & L & Avg \\
\midrule
0   & 91 & \textbf{98} & 93 & 75 & 89.3 \\
0.1 & \textbf{95} & 97 & \textbf{94} & \textbf{77} & \textbf{90.8} \\
0.2 & 92 & 97 & 92 & 76 & 89.3 \\
0.3 & 90 & 95 & 91 & 76 & 88.0 \\
\bottomrule
\end{tabular}
\caption{Reg. Loss Weight}
\label{tab:ablation-disp}
\end{subtable}
\hfill
\begin{subtable}[b]{0.32\linewidth}
\centering
\begin{tabular}{lccccc}
\toprule
Attn & S & O & G & L & Avg \\
\midrule
Full   & 90 & 95 & 93 & 76 & 88.5 \\
Causal & \textbf{95} & \textbf{97} & \textbf{94} & \textbf{77} & \textbf{90.8} \\
\bottomrule
\end{tabular}
\vspace{4mm}
\caption{Decoder Attention}
\label{tab:ablation-decoder}
\end{subtable}

\end{minipage}
}
\vspace{-4mm}
\end{table}

\paragraph{Number of Registers.}
We vary the number of registers $K$ to examine its influence on representation capacity. As shown in \cref{tab:ablation-register}, performance follows a clear non-monotonic trend. Increasing $K$ from 2 to 4 improves the average success rate, while further increasing $K$ leads to degradation. This indicates that simply enlarging the latent capacity does not monotonically improve performance; instead, there exists a moderate regime that balances expressiveness and optimization stability.

\paragraph{Regularization Loss Weight.}
We further analyze the effect of the regularization loss weight $\lambda_{\text{reg}}$, which controls the degree of latent separation in the learned action representations. 
As shown in \cref{tab:ablation-disp}, setting $\lambda_{\text{reg}}=0.1$ yields the highest average success rate. Removing the regularization term reduces average performance, while excessively large weights also degrade results. The optimal performance at $\lambda_{\text{reg}}=0.1$ suggests that moderate regularization is beneficial without overly constraining the latent space.

\paragraph{Decoder Attention Type.}
We further explore how the attention mechanism in the decoder influences overall performance. As shown in \cref{tab:ablation-decoder}, causal attention consistently outperforms full attention across all task suites, improving the average success rate by 2.25 points. This result indicates that explicitly enforcing temporal directionality during decoding remains beneficial even when actions are represented at the trajectory level.

\section{Conclusion}

We introduce CAT, a trajectory-level continuous action representation framework for robotic manipulation. 
CAT represents action trajectories within a fixed real-time interval using continuous latent tokens. 
This formulation enables policies to operate robustly under varying control frequencies.
Integrated with diffusion-based policy learning, CAT consistently improves policy performance across simulated benchmarks and real-robot manipulation tasks. 
These results highlight the effectiveness of trajectory-level continuous representations for visuomotor policy learning under varying control frequencies.

\clearpage

\renewcommand\thesection{\Alph{section}}
\setcounter{section}{0}

\renewcommand{\thefigure}{\Alph{figure}}
\setcounter{figure}{0}
\renewcommand{\thetable}{\Alph{table}}


\section{Implementation Details}
This section provides the architecture and training details used across all experiments for implementation clarity.
\subsection{Architecture} 
CAT is implemented as an 8-layer Transformer encoder paired with an 8-layer Transformer decoder, each with a hidden size of 512, 8 attention heads, and feed-forward networks with a 4$\times$ expansion ratio. The encoder uses full attention to aggregate information across the entire action chunk. The decoder adopts an asymmetric causal mask: decoder queries attend to all encoder outputs but only to past decoder positions, enforcing autoregressive latent reconstruction. RMSNorm~\cite{zhang2019root} is used in all Transformer blocks. The latent dimensionality is set to match the original action dimension.
\subsection{Training Pipeline}
Training follows a two-stage pipeline. In the first stage, CAT is trained to learn continuous latent representations of action trajectories. In the second stage, the pretrained CAT is integrated into a policy model, and the policy is trained for downstream control tasks while keeping CAT fixed. Unless otherwise specified, we use the Adam optimizer and fixed random seeds for reproducibility.

\subsection{Simulation Experiments}
\paragraph{\textbf{LIBERO.}}
For LIBERO simulation experiments, we use the cleaned and reformatted dataset released by the HuggingFace team following the preprocessing pipeline of OpenVLA~\cite{kim2024openvla}. The dataset removes no-op frames and failed episodes and contains 1,693 expert demonstrations converted to the LeRobot format. We use the same dataset version as SmolVLA.

CAT is trained on the preprocessed LIBERO dataset for 100k steps with batch size 64 and chunk size 16. The learning rate follows a cosine decay schedule from $1\mathrm{e}{-4}$ to $1\mathrm{e}{-5}$.
The pretrained CAT is integrated into the SmolVLA policy and trained on the same dataset for 100k steps with batch size 64 and chunk size 16. The learning rate follows a cosine decay from $1.5\mathrm{e}{-4}$ to $5\mathrm{e}{-6}$.

During evaluation, the action horizon is set to 8: the model predicts 16 future actions per step and executes the first 8. All experiments use a fixed random seed.

\paragraph{\textbf{MimicGen.}}
For MimicGen, we adopt the 8-task Robosuite subset and apply the same observation and action preprocessing as CARP~\cite{gong2025carp}. CAT is trained on trajectories from all eight tasks for 200 epochs with a learning rate of $1\mathrm{e}{-4}$. The pretrained CAT is then integrated into a flow-matching diffusion policy. The policy uses a 16-layer Transformer denoising network and is trained for 300 epochs on the 8-task dataset with batch size 1024 and learning rate $1\mathrm{e}{-4}$.

\paragraph{\textbf{RoboTwin 2.0 Platform.}}
We use RoboTwin~2.0 as the experimental platform and collect multi-frequency action trajectories for training CAT. For each task, trajectories are recorded at four control frequencies (10 Hz, 16.7 Hz, 25 Hz, and 50 Hz), with 50 episodes collected for each frequency.

CAT is trained on the full multi-frequency dataset using 1-second action chunks as input for 100k steps with batch size 1024 and learning rate $1\mathrm{e}{-4}$.
The pretrained CAT is then integrated into a SmolVLA policy following the training procedure described in~\cite{shukor2025smolvla}. Policies are trained separately for each frequency on eight manipulation tasks implemented in the RoboTwin~2.0 environment.

For evaluation, we report success rate as the primary metric. For a model trained at control frequency $f_s$, the action prediction horizon is set to $\lfloor f_s \rfloor$, and the first $\lfloor f_s/2 \rfloor$ actions are executed during rollout. The maximum episode length is defined using the timestep limit at 16.7Hz as a reference and scaled proportionally for other control frequencies.

\subsection{Real-World Experiments}
\begin{figure}[t]
\centering
\includegraphics[width=0.6\linewidth]{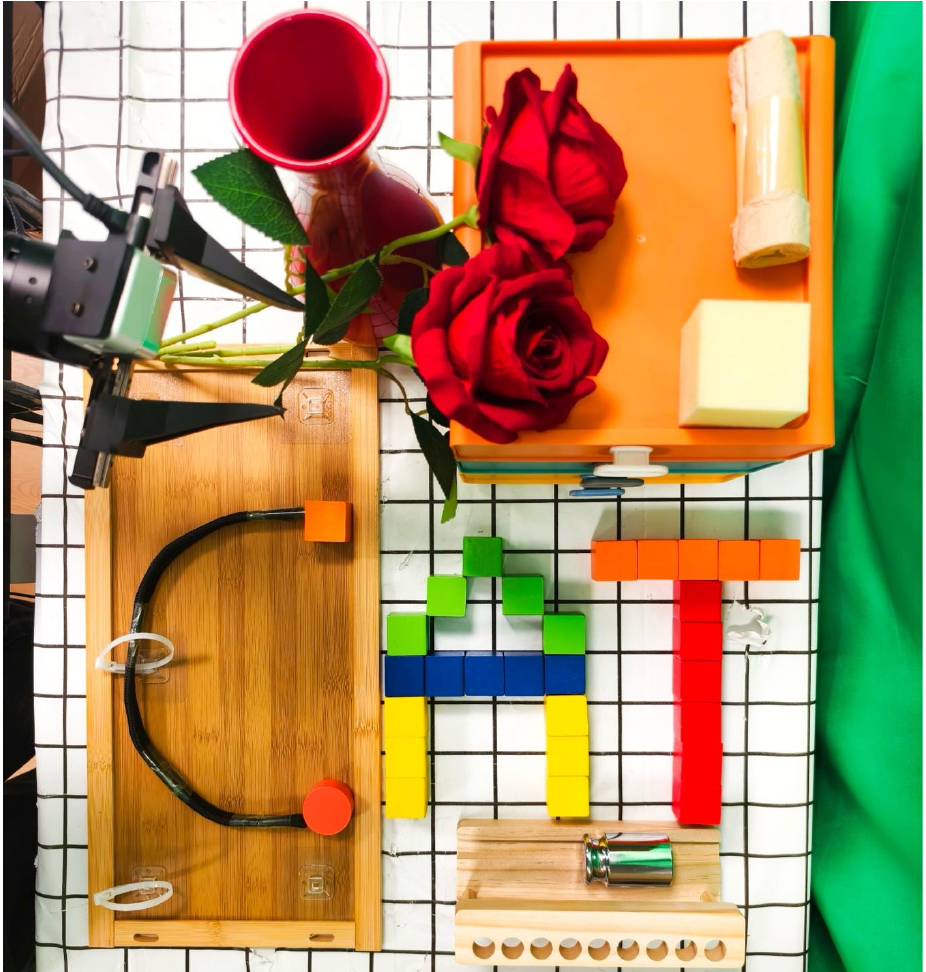}
\caption{\textbf{Real-world setup.} Objects used in our real-world experiments.}
\label{fig:real_world_setup}
\end{figure}

\cref{fig:real_world_setup} illustrates the objects used in our four real-world manipulation tasks. We first pretrain CAT on the OpenX dataset~\cite{collaboration2023open}. Training runs for 400k steps with batch size 1024 using 1-second action chunks and a learning rate of $1\mathrm{e}{-4}$. The CAT is then fine-tuned separately for each real-world task on tele-operated data collected at 20 Hz for 1k steps using 16-step action chunks. For policy learning, SmolVLA-CAT is trained for 180k steps with learning rate $2\mathrm{e}{-4}$, batch size 64, and action chunk size 16. The SmolVLA baseline uses the same training setup.

\section{Qualitative Results}

We provide qualitative visualizations of successful rollouts across all benchmarks. 
In each figure, a row corresponds to a single task rollout, where the leftmost frame 
shows the initial state and the rightmost frame shows the final state.

\begin{figure*}[t]
\centering

\begin{subfigure}{\linewidth}
    \centering
    \includegraphics[width=\linewidth]{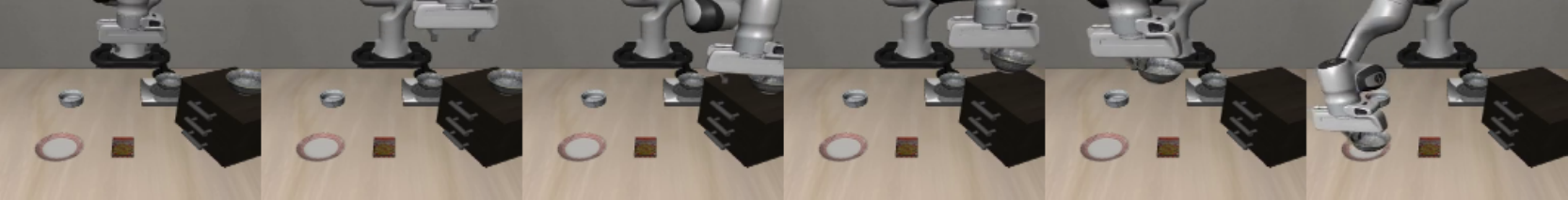}
    \caption{Spatial: pick up the black bowl on the wooden cabinet and place it on the plate.}
\end{subfigure}

\begin{subfigure}{\linewidth}
    \centering
    \includegraphics[width=\linewidth]{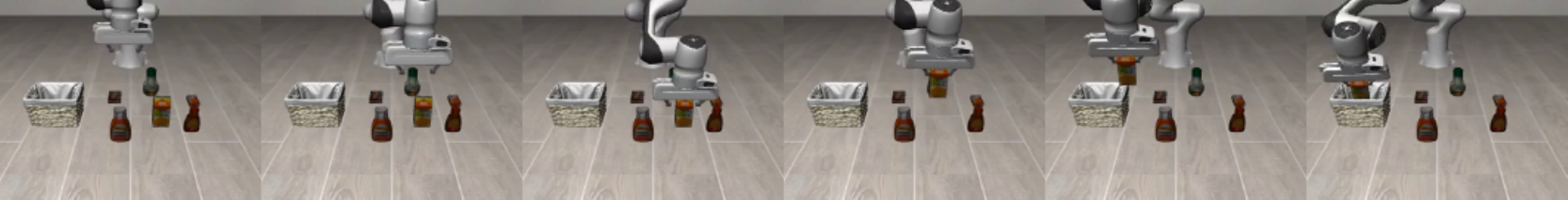}
    \caption{Object: pick up the orange juice and place it in the basket.}
\end{subfigure}

\begin{subfigure}{\linewidth}
    \centering
    \includegraphics[width=\linewidth]{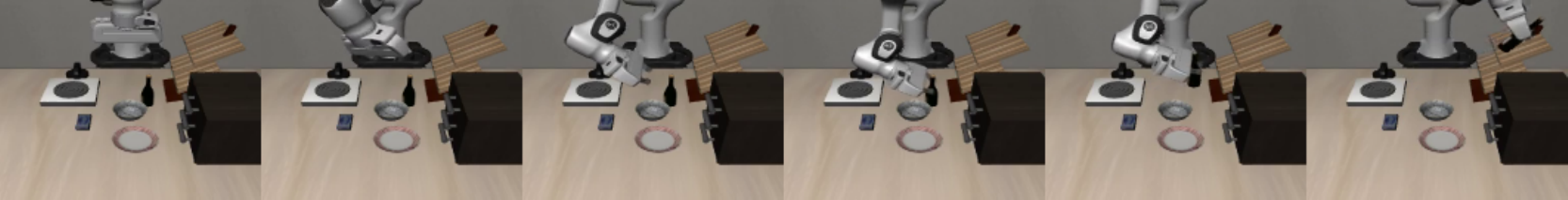}
    \caption{Goal: put the wine bottle on the rack.}
\end{subfigure}

\begin{subfigure}{\linewidth}
    \centering
    \includegraphics[width=\linewidth]{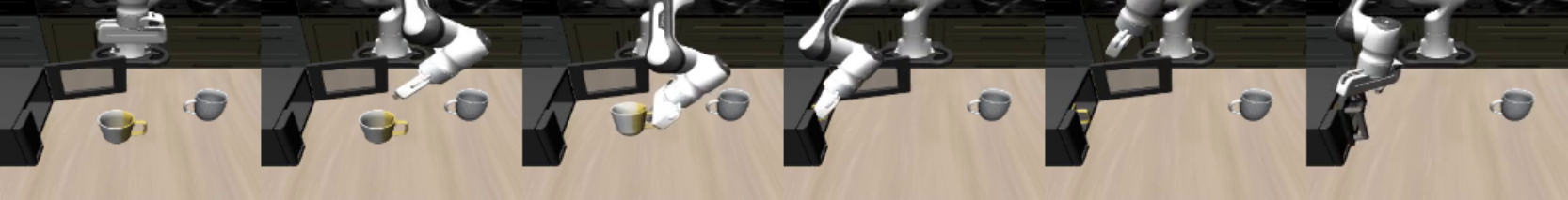}
    \caption{Long: put the yellow and white mug in the microwave and close it.}
\end{subfigure}

\caption{
\textbf{Qualitative results on LIBERO tasks.}
Representative examples include tasks from Spatial, Object, Goal and Long.
}
\label{fig:libero_vis}
\end{figure*}

\paragraph{\textbf{LIBERO.}}
As shown in \cref{fig:libero_vis}, we present rollout frames from four task suites:
\emph{Spatial}, \emph{Object}, \emph{Goal}, and \emph{Long}. 
Each row corresponds to one task suite, with several frames uniformly sampled 
from a successful episode.

\begin{figure*}[t]
\centering

\begin{subfigure}{\linewidth}
    \centering
    \includegraphics[width=\linewidth]{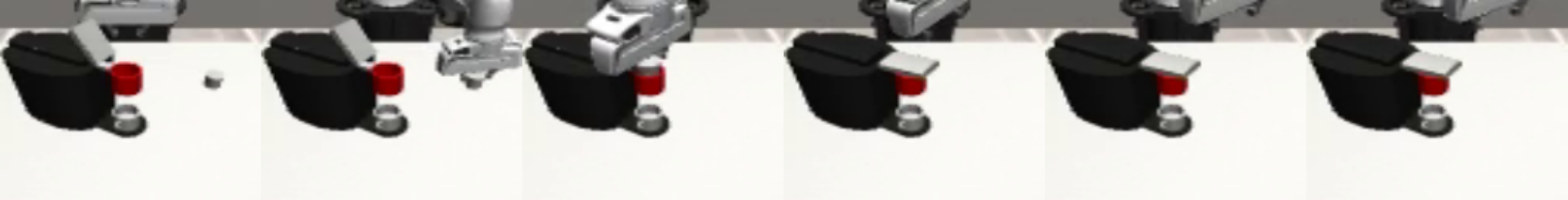}
    \caption{Coffee: pick a coffee pod, insert into the coffee machine, and close the machine hinge.}
\end{subfigure}

\begin{subfigure}{\linewidth}
    \centering
    \includegraphics[width=\linewidth]{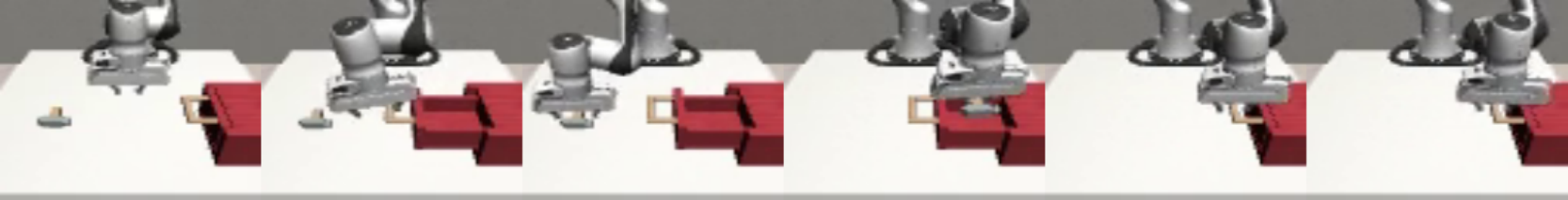}
    \caption{Hammer: open the drawer, pick the hammer, and place into the drawer.}
\end{subfigure}

\begin{subfigure}{\linewidth}
    \centering
    \includegraphics[width=\linewidth]{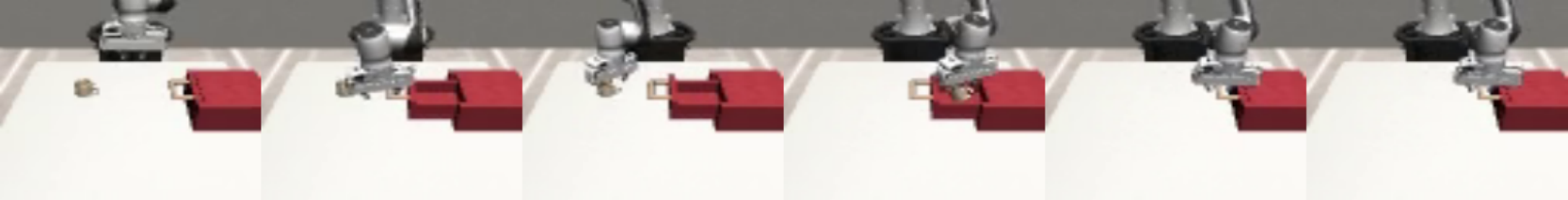}
    \caption{Mug: open the drawer, pick the mug, and place into the drawer.}
\end{subfigure}

\begin{subfigure}{\linewidth}
    \centering
    \includegraphics[width=\linewidth]{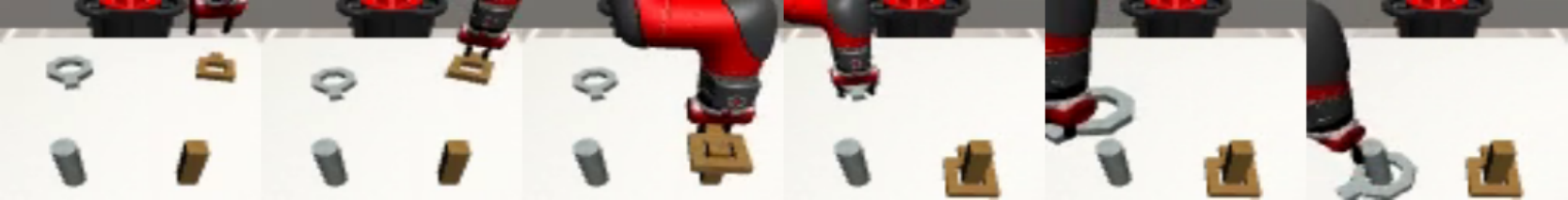}
    \caption{Nut: grasp the square nut, insert it onto the square peg, grasp the circle nut, and insert it onto the circle peg}
\end{subfigure}

\begin{subfigure}{\linewidth}
    \centering
    \includegraphics[width=\linewidth]{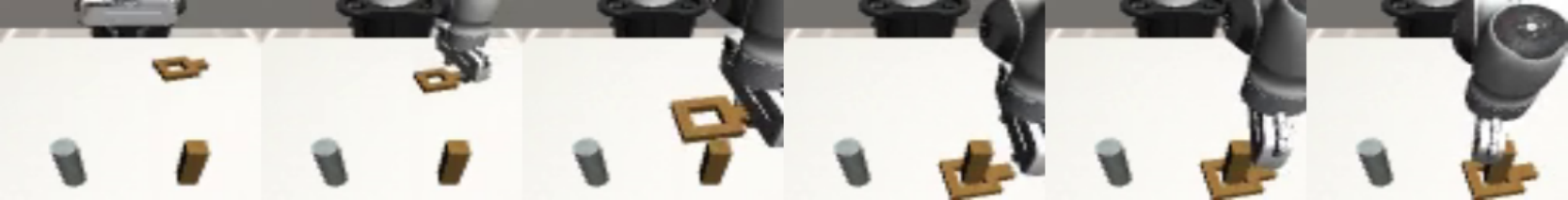}
    \caption{Square: grasp the nut and then insert it onto the peg.}
\end{subfigure}

\begin{subfigure}{\linewidth}
    \centering
    \includegraphics[width=\linewidth]{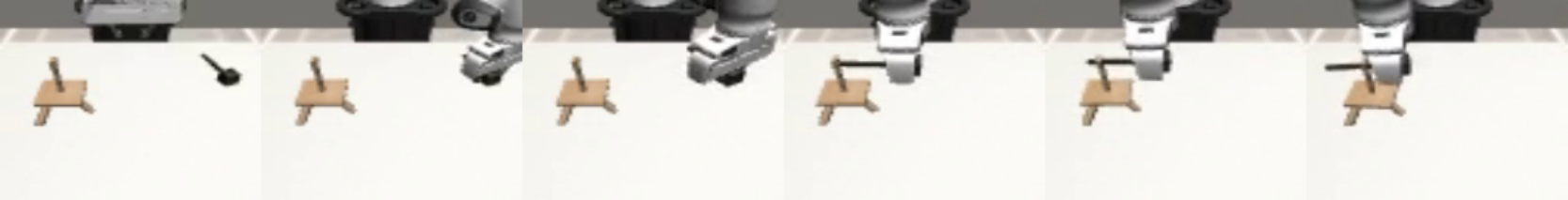}
    \caption{Threading: insert a thin shaft through a tight ring.}
\end{subfigure}

\begin{subfigure}{\linewidth}
    \centering
    \includegraphics[width=\linewidth]{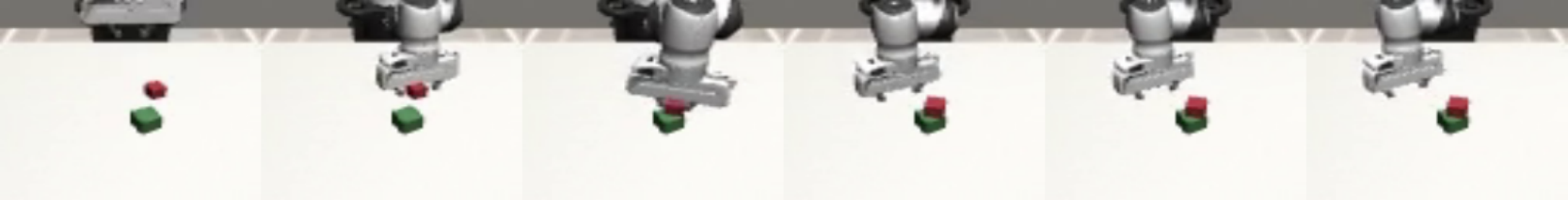}
    \caption{Stack: stack two blocks in red-green order.}
\end{subfigure}

\begin{subfigure}{\linewidth}
    \centering
    \includegraphics[width=\linewidth]{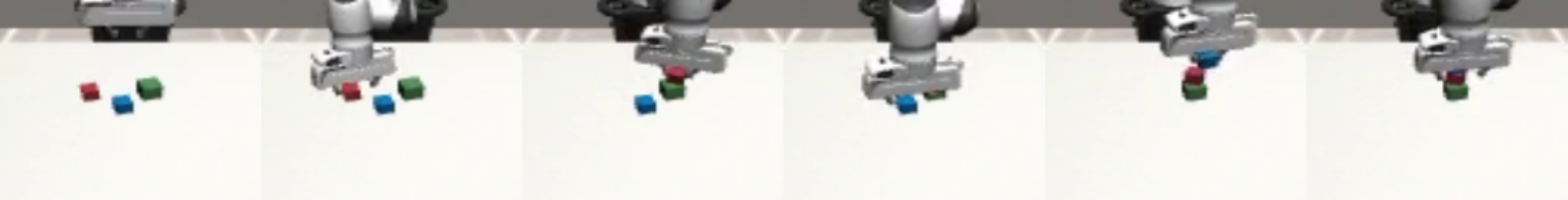}
    \caption{Stack3: stack three blocks in blue-red-green order.}
\end{subfigure}

\caption{
\textbf{Qualitative results on MimicGen tasks.} Representative examples include tasks from Coffee, Hammer, Mug, Square, Threading, Stack, Stack3.
}
\label{fig:mimicgen}

\end{figure*}

\paragraph{\textbf{MimicGen.}}
As shown in \cref{fig:mimicgen}, we show rollout frames from eight task suites:
\emph{Coffee}, \emph{Hammer}, \emph{Mug}, \emph{Nut}, \emph{Square}, 
\emph{Threading}, \emph{Stack}, and \emph{Stack3}. 
Each row corresponds to one task suite, where frames are sampled 
from a successful episode.

\begin{figure*}[t]
\centering

\begin{subfigure}{\linewidth}
    \centering
    \includegraphics[width=\linewidth]{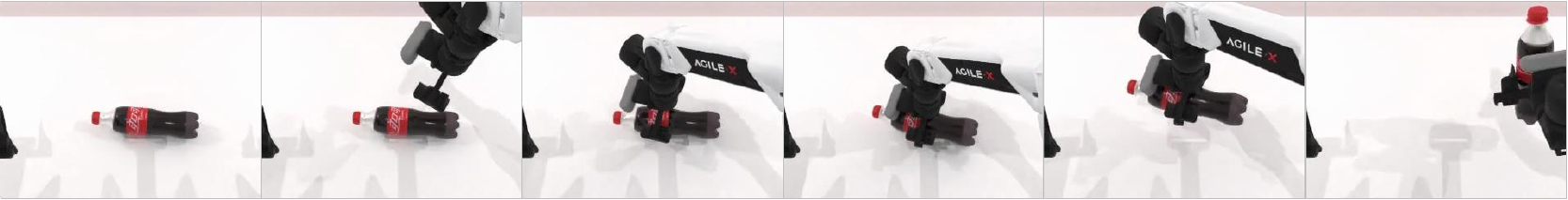}
    \caption{Adjust Bottle: pick up the bottle on the table headup with the correct arm.}
\end{subfigure}

\begin{subfigure}{\linewidth}
    \centering
    \includegraphics[width=\linewidth]{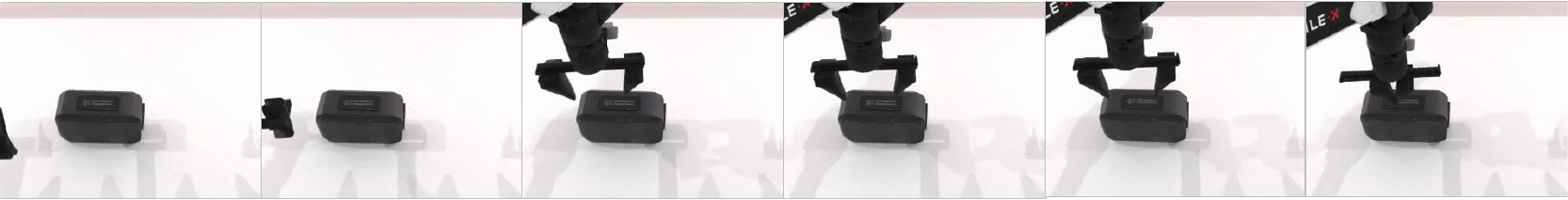}
    \caption{Click Alarmclock: click the alarm clock's center of the top side button on the table.}
\end{subfigure}

\begin{subfigure}{\linewidth}
    \centering
    \includegraphics[width=\linewidth]{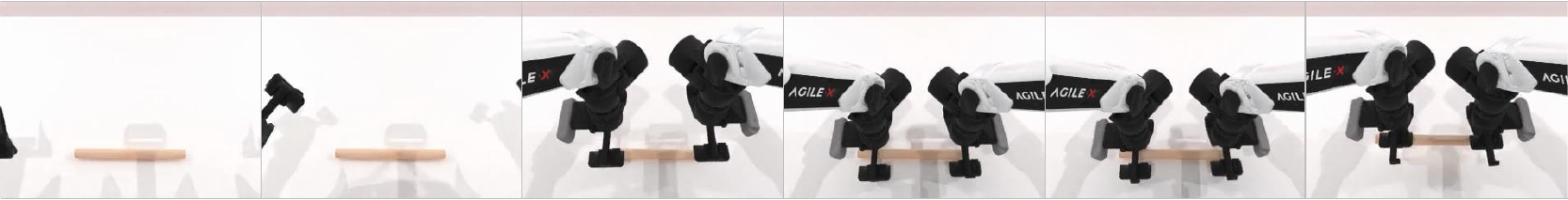}
    \caption{Grab Roller: use both arms to grab the roller on the table.}
\end{subfigure}

\begin{subfigure}{\linewidth}
    \centering
    \includegraphics[width=\linewidth]{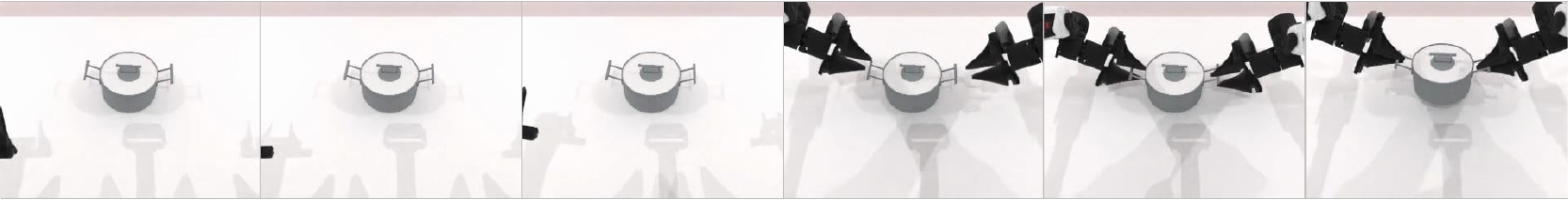}
    \caption{Lift Pot: use arms to lift the pot.}
\end{subfigure}

\begin{subfigure}{\linewidth}
    \centering
    \includegraphics[width=\linewidth]{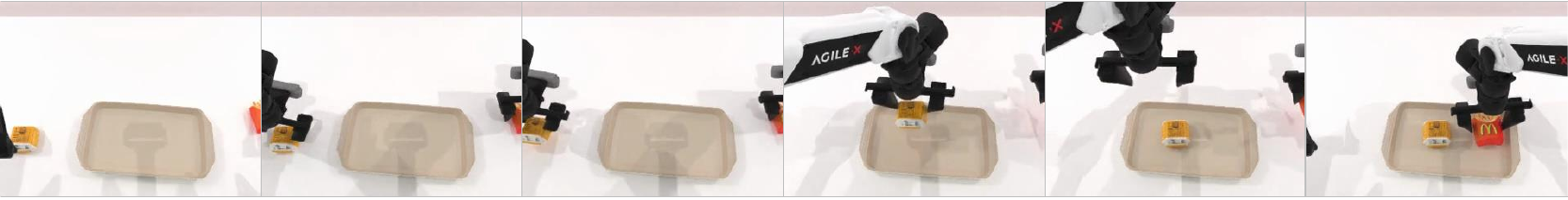}
    \caption{Place Burger Fries: use dual arm to pick the hamburg and frenchfries and put them onto the tray.}
\end{subfigure}

\begin{subfigure}{\linewidth}
    \centering
    \includegraphics[width=\linewidth]{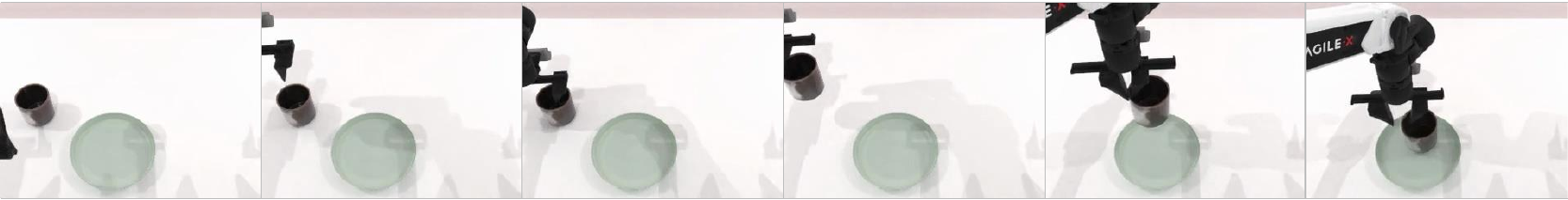}
    \caption{Place Container Plate: place the container onto the plate.}
\end{subfigure}

\begin{subfigure}{\linewidth}
    \centering
    \includegraphics[width=\linewidth]{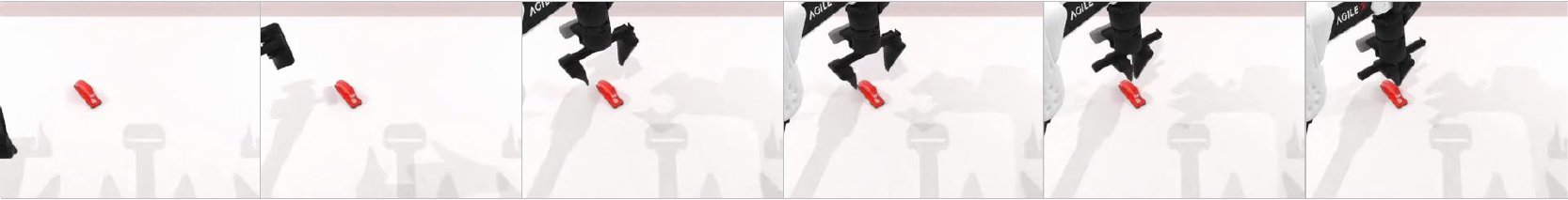}
    \caption{Press Stapler: use one arm to press the stapler.}
\end{subfigure}

\begin{subfigure}{\linewidth}
    \centering
    \includegraphics[width=\linewidth]{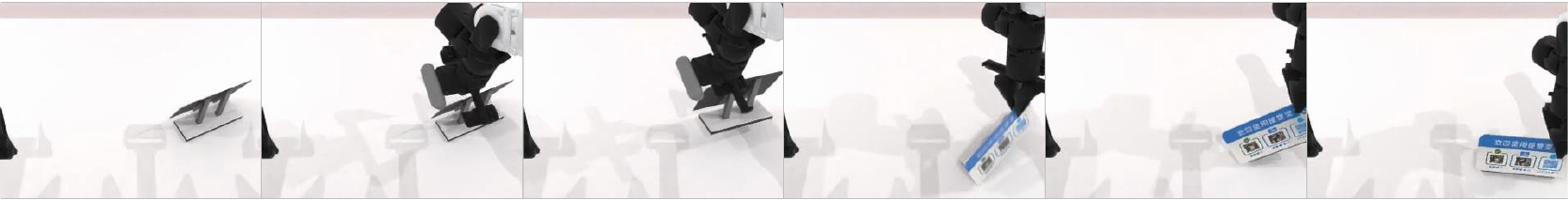}
    \caption{Rotate QRcode: use arm to catch the qrcode board on the table, pick it up and rotate to let the qrcode face towards the robot.}
\end{subfigure}

\caption{
\textbf{Qualitative results on RoboTwin~2.0 tasks at a test-time control frequency of 16.7Hz.}
Representative examples include tasks from Adjust Bottle, Click Alarmclock, Grab Roller, Lift Pot, Place Burger Fries, Place Container Plate, Press Stapler, and Rotate QRcode.
}
\label{fig:robotwin}
\end{figure*}

\paragraph{\textbf{RoboTwin 2.0 Platform.}}
As shown in \cref{fig:robotwin}, we visualize rollouts from eight manipulation tasks:
\emph{Adjust Bottle}, \emph{Click Alarmclock}, \emph{Grab Roller}, \emph{Lift Pot},
\emph{Place Burger Fries}, \emph{Place Container Plate}, \emph{Press Stapler}, 
and \emph{Rotate QRcode}. 
All rollouts are evaluated at a test-time control frequency of 16.7Hz.

\clearpage
\section*{Acknowledgements}
This work was supported by National Natural Science Foundation of China (No.62576109), Scientific and Technological innovation action plan of  Shanghai Science and Technology Committee (No.25511104402).

%
%
\bibliographystyle{splncs04}
\bibliography{main}
\end{document}